# A Simple Algorithm for Semi-supervised Learning with Improved Generalization Error Bound

**Ming Ji**[∗‡] MINGJI1@ILLINOIS.EDU
**Tianbao Yang**[∗†] YANGTIA1@MSU.EDU
**Binbin Lin**[♮] BINBINLIN@ZJU.EDU.CN
**Rong Jin**[†] RONGJIN@CSE.MSU.EDU
**Jiawei Han**[‡] HANJ@ILLINOIS.EDU

[‡]Department of Computer Science, University of Illinois at Urbana-Champaign, Urbana, IL, 61801, USA
[†]Department of Computer Science and Engineering, Michigan State University, East Lansing, MI, 48824, USA
[♮]State Key Lab of CAD&CG, College of Computer Science, Zhejiang University, Hangzhou, 310058, China
[∗]Equal contribution

## Abstract

In this work, we develop a simple algorithm for semi-supervised regression. The key idea is to use the top eigenfunctions of integral operator derived from both labeled and unlabeled examples as the basis functions and learn the prediction function by a simple linear regression. We show that under appropriate assumptions about the integral operator, this approach is able to achieve an improved regression error bound better than existing bounds of supervised learning. We also verify the effectiveness of the proposed algorithm by an empirical study.

## 1. Introduction

Although numerous algorithms have been developed for semi-supervised learning (Zhu (2008) and references therein), most of them do not have theoretical guarantee on improving the generalization performance of supervised learning. A number of theories have been proposed for semi-supervised learning, and most of them are based on one of the two assumptions: (1) the *cluster assumption* (Seeger, 2001; Rigollet, 2007; Lafferty & Wasserman, 2007; Singh et al., 2008; Sinha & Belkin, 2009) which assumes that two data points should have the same class label or similar values if they are connected by a path passing through a high density region; (2) the *manifold assumption* (Lafferty & Wasserman, 2007; Niyogi, 2008) which states that the prediction function lives in a low dimensional manifold of the marginal distribution $P_{\mathcal{X}}$.

It has been pointed out by several studies (Lafferty & Wasserman, 2007; Nadler et al., 2009) that the manifold assumption by itself is insufficient to reduce the generalization error bound of supervised learning. However, on the other hand, it was found in (Niyogi, 2008) that for certain learning problems, no supervised learner can learn effectively, while a manifold based learner (that knows the manifold or learns it from unlabeled examples) can learn well with relatively few labeled examples. Compared to the manifold assumption, theoretical results based on cluster assumption appear to be more encouraging. In the early studies (Castelli & Cover, 1995; 1996), the authors show that under the assumption that the marginal distribution $P_{\mathcal{X}}$ is a mixture of class conditional distributions, the generalization error will be reduced exponentially in the number of labeled examples if the mixture is identifiable. Rigollet (2007) defines the cluster assumption in terms of density level sets, and shows a similar exponential convergence rate given a sufficiently large number of unlabeled examples. Furthermore, Singh et al. (2008) show that the mixture components can be identified if $P_{\mathcal{X}}$ is a mixture of a finite number of smooth density functions and the separation/overlap between different mixture components is significantly large. Despite the encouraging results, one major problem of the cluster assumption is that it is difficult to be verified given a limited number of labeled examples. In addition, the learning algorithms suggested in (Rigollet, 2007; Singh et al., 2008; Zhang & Ando, 2005) are difficult to implement efficiently even if the cluster assumption holds, making them unpractical

Appearing in *Proceedings of the 29$^{th}$ International Conference on Machine Learning*, Edinburgh, Scotland, UK, 2012.

for real-world problems.

In this work, we aim to develop a *simple* algorithm for semi-supervised learning that on one hand is easy to implement, and on the other hand is guaranteed to improve the generalization performance of supervised learning under appropriate assumptions. The main idea of the proposed algorithm is to estimate the top eigenfunctions of the integral operator from the both labeled and unlabeled examples, and learn from the labeled examples the best prediction function in the subspace spanned by the estimated eigenfunctions. Unlike the previous studies of exploring eigenfunctions for semi-supervised learning (Fergus et al., 2009; Sinha & Belkin, 2009), we show that under appropriate assumptions, the proposed algorithm achieves a better generalization error bound than supervised learning algorithms.

To derive the generalization error bound, we make a different set of assumptions from previous studies. First, we assume a skewed eigenvalue distribution and bounded eigenfunctions of the integral operator. The assumption of skewed eigenvalue distributions has been verified and used in multiple studies of kernel learning (Koltchinskii, 2011; Steinwart et al., 2006; Minh, 2010; Zhang & Ando, 2005), while the assumption of bounded eigenvectors was mostly found in the study of compressive sensing (Candès & Tao, 2006). Second, we assume that a sufficient number of labeled examples are available, which is also used by the other analysis of semi-supervised learning (Rigollet, 2007). It is the combination of these assumptions that allow us to derive better generalization error bound for semi-supervised learning.

The rest of the paper is arranged as follows. Section 2 presents the proposed algorithm and verifies its effectiveness by an empirical study. Section 3 shows the improved generalization error bound for the proposed semi-supervised learning, and Section 4 outlines the proofs. Section 5 concludes with future work.

## 2. Algorithm and Empirical Validation

Let $\mathcal{X}$ be a compact domain or a manifold in the Euclidean space $\mathbb{R}^d$. Let $\mathcal{D} = \{\mathbf{x}_i, i = 1, \ldots, N \,|\mathbf{x}_i \in \mathcal{X}\}$ be a collection of training examples. We randomly select $n$ examples from $\mathcal{D}$ for labeling. Without loss of generality, we assume that the first $n$ examples are labeled by $\mathbf{y}_l = (y_1, \ldots, y_n)^\top \in \mathbb{R}^n$. We denote by $\mathbf{y} = (y_1, \ldots, y_N)^\top \in \mathbb{R}^N$ the true labels for all the examples in $\mathcal{D}$. In this study, we assume $y = f(\mathbf{x})$ is decided by an unknown deterministic function $f(\mathbf{x})$. Our goal is to learn an accurate prediction function by exploiting both labeled and unlabeled examples. Below we first present our algorithm and then verify its empirical performance by comparing to the state-of-the-art algorithms for supervised and semi-supervised learning.

**Algorithm 1** A Simple Algorithm for Semi-supervised Learning

1: **Input**
- $\mathcal{D} = \{\mathbf{x}_1, \ldots, \mathbf{x}_N\}$: labeled and unlabeled examples
- $\mathbf{y}_l = (y_1, \ldots, y_n)^\top$: labels for the first $n$ examples in $\mathcal{D}$
- $s$: the number of eigenfunctions to be used

2: Compute $(\widehat{\phi}_i, \widehat{\lambda}_i), i = 1, \ldots, s$, the first $s$ eigenfunctions and eigenvalues for the integral operator $\widehat{L}_N$ defined in (4).

3: Compute the prediction $\widehat{g}(\mathbf{x})$ in (5), where $\gamma^* = (\gamma_1^*, \ldots, \gamma_s^*)^\top$ is given by solving the following regression problem

$$\gamma^* = \arg\min_{\gamma \in \mathbb{R}^s} \sum_{i=1}^n \left( \sum_{j=1}^s \gamma_j \widehat{\phi}_j(\mathbf{x}_i) - y_i \right)^2 \quad (1)$$

4: **Output** prediction function $\widehat{g}(\cdot)$

### 2.1. A Simple algorithm for Semi-Supervised Learning

Let $\kappa(\cdot, \cdot) : \mathcal{X} \times \mathcal{X} \to \mathbb{R}$ be a Mercer kernel, and let $\mathcal{H}_\kappa$ be a Reproducing Kernel Hilbert space (RKHS) of functions $\mathcal{X} \to \mathbb{R}$ endowed with kernel $\kappa(\cdot, \cdot)$. We assume that $\kappa$ is a bounded function, i.e., $|\kappa(\mathbf{x}, \mathbf{x})| \leq 1, \forall \mathbf{x} \in \mathcal{X}$. Similar to most semi-supervised learning algorithms, in order to effectively exploit the unlabeled data, we need to relate the prediction function $f(\mathbf{x})$ to the unlabeled examples (or the marginal distribution $P_\mathcal{X}$). To this end, we assume there exists an accurate prediction function $g(\mathbf{x}) \in \mathcal{H}_\kappa$ with $\|g\|_{\mathcal{H}_\kappa} \leq R$. More specifically, we define

$$\varepsilon^2 = \min_{h \in \mathcal{H}_\kappa, \|h\|_{\mathcal{H}_\kappa} \leq R} \mathrm{E}_\mathbf{x}[(f(\mathbf{x}) - h(\mathbf{x}))^2], \quad (2)$$

$$g(\mathbf{x}) = \arg\min_{h \in \mathcal{H}_\kappa, \|h\|_{\mathcal{H}_\kappa} \leq R} \mathrm{E}_\mathbf{x}[(f(\mathbf{x}) - h(\mathbf{x}))^2]. \quad (3)$$

Our basic assumption (**A0**) is that the regression error $\varepsilon^2 \ll R^2$ is small, and the maximum regression error of $g(\mathbf{x})$ for any $\mathbf{x} \in \mathcal{X}$ is also small, i.e.,

$$\sup_{x \in \mathcal{X}} (f(\mathbf{x}) - g(\mathbf{x}))^2 \triangleq \varepsilon_{\max}^2 = O(n\varepsilon^2 / \ln N).$$

To present our algorithm, we define an integral oper-ator over the examples in $\mathcal{D}$:

$$\widehat{L}_N(f)(\cdot) = \frac{1}{N}\sum_{i=1}^{N}\kappa(\mathbf{x}_i, \cdot)f(\mathbf{x}_i), \quad (4)$$

where $f \in \mathcal{H}_\kappa$. Let $(\widehat{\phi}_i(\mathbf{x}), \widehat{\lambda}_i), i = 1, 2, \ldots, N$ be the eigenfunctions and eigenvalues of $\widehat{L}_N$ ranked in the descending order of eigenvalues, where $\langle \widehat{\phi}_i(\cdot), \widehat{\phi}_j(\cdot)\rangle_{\mathcal{H}_\kappa} = \delta(i, j)$ for any $1 \leq i, j \leq N$. According to (Guo & Zhou, 2011), the prediction function $g(\mathbf{x})$ can be well approximated by a function in the subspace spanned by the top eigenfunctions of $\widehat{L}_N$. Hence, we propose to learn a target prediction function $\widehat{g}(\mathbf{x})$ as a linear combination of the first $s$ eigenfunctions, i.e.,

$$\widehat{g}(\mathbf{x}) = \sum_{j=1}^{s}\gamma_j^*\widehat{\phi}_j(\mathbf{x}), \quad (5)$$

where $s$ is a parameter that needs to be determined empirically. Coefficients $\{\gamma_i^*\}_{i=1}^{s}$ in (5) are learned through a simple regression by minimizing the squared error of the labeled examples as shown in (1). Algorithm 1 shows the basic steps of the proposed algorithm.

**Implementation** In step 2 of Algorithm 1, we need to compute the eigenvalues and eigenfunctions of $\widehat{L}_N$, which is given as follows (Smale & Zhou, 2009). Let $K = [\kappa(\mathbf{x}_i, \mathbf{x}_j)]_{N\times N}$ be the kernel matrix for the examples in $\mathcal{D}$, and let $\{(\mathbf{v}^i, \sigma_i)\}_{i=1}^{s}$ be the first $s$ eigenvectors and eigenvalues of $K$. Then, the eigenvalues and eigenfunctions of $\widehat{L}_N$ are given by

$$\widehat{\lambda}_i = \frac{\sigma_i}{N},\ \widehat{\phi}_i(\cdot) = \frac{1}{\sqrt{\sigma_i}}\sum_{j=1}^{N} v_j^i\kappa(\mathbf{x}_j, \cdot),\ i = 1, \ldots, s,$$

where $v_j^i$ is the $j$-th element of vector $\mathbf{v}^i$. Finally, in step 3 of Algorithm 1, we need to compute the optimal coefficient $\gamma^*$, which, according to (Bishop, 2006), is given by

$$\gamma^* = D^{1/2}[V^\top K_B K_B^\top V]^{-1}V^\top K_B\mathbf{y}_l,$$

where $D = \mathrm{diag}(\sigma_1, \ldots, \sigma_s)$, $K_B = [\kappa(\mathbf{x}_i, \mathbf{x}_j)]_{N\times n}$ includes the kernel similarity between all the examples in $\mathcal{D}$ and labeled examples, and $V = (\mathbf{v}^1, \ldots, \mathbf{v}^s)$.

### 2.2. Empirical study

Three real-world data sets, i.e., insurance, wine, and temperature [1], are used in our empirical study. The statistics of these datasets are given in Table 1. The first two datasets are from the UC Irvine Machine Learning Repository (Frank & Asuncion, 2010), while the task of the last dataset is to predict the temperature based on the coordinates (latitude, longitude) on the earth surface. All three datasets are designed for regression tasks with real-valued outputs. We choose these three datasets because they fit in with our assumptions that will be elaborated in section 3.2.

*Table 1.* Statistics of datasets

| Name | #Objects | #Features |
|---|---|---|
| insurance | 9,822 | 85 |
| wine | 4,898 | 11 |
| temperature | 9,504 | 2 |

We randomly choose 90% of the data for training, and use the rest 10% for testing. We randomly select $2\%, 3\%, \ldots, 9\%$ of the entire dataset as labeled examples. We evaluate the performance by measuring the regression error of the testing data. Each experiment is repeated ten times and the regression errors averaged over the ten trials are reported. Two supervised regression algorithms, i.e., Kernel Ridge Regression (**KRR**) (Saunders et al., 1998) and Support Vector Regression (**SVR**) (Drucker et al., 1996), and a state-of-the-art algorithm for semi-supervised regression, i.e., Laplacian Regularized Least Squares (**LapRLS**) (Belkin et al., 2006), are used as the baselines. We did not include other baseline algorithms for semi-supervised learning because Laplacian regularization yields the state-of-the-art performance of semi-supervised learning. More importantly, our goal is to verify that the proposed algorithm can effectively improve the generalization performance of supervised learning. We refer to the proposed algorithm as Simple Semi-Supervised Learning, or **SSSL** for short. A RBF kernel function is used for all algorithms, and all the parameters are chosen by cross validation.

Tables 2-4 show the regression errors for the three datasets, respectively. First, as we expected, the performance of all learning algorithms improves as the number of labeled examples increases. It is also not surprising to see that the two semi-supervised learning algorithms perform better than the two supervised learning algorithms. Second, the proposed algorithm (SSSL) outperforms the baseline semi-supervised learning algorithm for almost all the cases, indicating that it is effective for semi-supervised learning. Note that SVR does not perform well on the temperature dataset since this dataset has a perfect manifold structure (the earth surface is a sphere), and SVR fails to capture the manifold structure when the percentage of labeled data is very small.

[1] http://www.remss.com/msu

*Table 2.* Regression error for the insurance data set (mean ± std)

| % labeled data | 2% | 3% | 4% | 5% | 6% | 7% | 8% | 9% |
|---|---|---|---|---|---|---|---|---|
| KRR | 0.0804 ±0.0084 | 0.0778 ±0.0088 | 0.0779 ±0.0125 | 0.0747 ±0.0099 | 0.0739 ±0.0100 | 0.0711 ±0.0071 | 0.0672 ±0.0065 | 0.0675 ±0.0065 |
| SVR | 0.0546 ±0.0038 | 0.0546 ±0.0040 | 0.0546 ±0.0040 | 0.0549 ±0.0039 | 0.0550 ±0.0038 | 0.0548 ±0.0040 | 0.0549 ±0.0041 | 0.0550 ±0.0040 |
| LapRLS | 0.0550 ±0.0044 | 0.0563 ±0.0060 | 0.0580 ±0.0068 | 0.0559 ±0.0048 | 0.0564 ±0.0052 | 0.0547 ±0.0039 | 0.0538 ±0.0053 | 0.0543 ±0.0046 |
| SSSL | **0.0544 ±0.0051** | **0.0527 ±0.0038** | **0.0527 ±0.0041** | **0.0526 ±0.0042** | **0.0523 ±0.0038** | **0.0518 ±0.0041** | **0.0518 ±0.0040** | **0.0517 ±0.0040** |

*Table 3.* Regression error for the wine dataset (mean ± std)

| % labeled data | 2% | 3% | 4% | 5% | 6% | 7% | 8% | 9% |
|---|---|---|---|---|---|---|---|---|
| KRR | 0.931 ±0.104 | 0.927 ±0.1289 | 0.799 ±0.102 | 0.759 ±0.149 | 0.714 ±0.056 | 0.681 ±0.086 | 0.650 ±0.086 | 0.668 ±0.079 |
| SVR | 0.669 ±0.038 | 0.642 ±0.037 | 0.656 ±0.035 | 0.613 ±0.023 | 0.613 ±0.029 | 0.606 ±0.017 | 0.600 ±0.020 | 0.592 ±0.028 |
| LapRLS | 0.682 ±0.038 | 0.653 ±0.042 | 0.650 ±0.035 | 0.613 ±0.025 | 0.611 ±0.022 | 0.597 ±0.023 | 0.592 ±0.017 | **0.580 ±0.022** |
| SSSL | **0.612 ±0.027** | **0.606 ±0.029** | **0.599 ±0.029** | **0.593 ±0.030** | **0.587 ±0.027** | **0.582 ±0.026** | **0.584 ±0.033** | 0.581 ±0.029 |

*Table 4.* Regression error for the temperature dataset (mean ± std)

| % labeled data | 2% | 3% | 4% | 5% | 6% | 7% | 8% | 9% |
|---|---|---|---|---|---|---|---|---|
| KRR | 8.69 ±0.84 | 7.61 ±0.63 | 7.16 ±0.33 | 7.04 ±0.45 | 6.81 ±0.50 | 6.61 ±0.46 | 6.46 ±0.33 | 6.29 ±0.36 |
| SVR | 82.3 ±2.8 | 79.0 ±3.2 | 74.0 ±2.5 | 72.5 ±2.1 | 68.1 ±1.9 | 63.9 ±2.5 | 61.5 ±2.5 | 59.1 ±2.9 |
| LapRLS | 6.78 ±0.62 | 6.05 ±0.36 | 5.88 ±0.25 | 5.76 ±0.28 | 5.73 ±0.32 | 5.63 ±0.38 | 5.54 ±0.28 | 5.42 ±0.28 |
| SSSL | **3.52 ±0.57** | **2.73 ±0.31** | **2.55 ±0.17** | **2.55 ±0.17** | **2.54 ±0.11** | **2.47 ±0.14** | **2.40 ±0.16** | **2.35 ±0.11** |

## 3. Generalization Error Bounds

To analyze the generalization performance of the proposed algorithm, we first consider the simple scenario where we have access to an infinite number of unlabeled examples (i.e., the marginal distribution $P_{\mathcal{X}}$). We then present the generalization error bound for a finite number of unlabeled examples. Detailed analysis can be found in Section 4.

### 3.1. Generalization error for an infinite number of unlabeled examples

Given the marginal distribution $P_{\mathcal{X}}$, we define an integral operator $L$ as $L(f)(\cdot) = \mathrm{E}_{\mathbf{x}}[\kappa(\mathbf{x}, \cdot)f(\mathbf{x})]$. We denote by $\{(\phi_i(\cdot), \lambda_i), i = 1, 2, \ldots\}$ the eigenfunctions and eigenvalues of $L$ ranked in the descending order of the eigenvalues, where the eigenfunctions are normalized according to the distribution, i.e., $\int_{\mathbf{x}\in\mathcal{X}} \phi_i(\mathbf{x})\phi_j(\mathbf{x})dP_{\mathcal{X}} = \delta_{ij}$. We note that $\widehat{L}_N$, defined in (4), is the empirical version of $L$, and $\|L - \widehat{L}_N\|_{HS}$ approaches to zero as the number of examples goes to infinity, where $\|\cdot\|_{HS}$ is Hilbert Schmidt norm of a linear operator (Smale & Zhou, 2009).

In order to achieve a better generalization error bound for the proposed semi-supervised learning algorithm, we make the following assumptions about eigenvalues and eigenfunctions:

- **A1** *Skewed eigenvalue distribution.* Similar to many studies (Koltchinskii & Yuan, 2010; Steinwart et al., 2006; Minh, 2010), we assume the eigenvalues follow a power law distribution, i.e., there exists a small constant $a > 0$ and a power index $p > 2$, such that
$$\lambda_k \leq a^2 k^{-p}, k = 1, 2, \ldots.$$
- **A2** *Bounded eigenfunctions.* There exists a small constant $C$ such that $\max_{\mathbf{x}\in\mathcal{X}} \max_i |\phi_i(\mathbf{x})| \leq C$. This is similar to the incoherence condition specified in compressive sensing (Candès & Tao, 2006).
- **A3** *Sufficient number of labeled examples.* We require the number of labeled examples to be larger than $n_0$ which is defined as
$$n_0 = 64C^2 \ln^2(2N^3) \left(\frac{Ra}{\varepsilon}\right)^{4/(p-1)}, \qquad (6)$$

where $N > 0$ is some large number that corresponds to the number of unlabeled examples when we come to the case of finite samples.

**Remark 1** Assumption **(A1)** ensures that the target function can be approximated, with a small error, by a function in the subspace spanned by the top eigenfunctions of $L$. This is the foundation behind Algorithm 1.

**Remark 2** Assumptions **(A2)** and **(A3)** are introduced to ensure that *all* the coefficients $\{\gamma_i^*\}_{i=1}^s$ in (5) can be estimated accurately. More specifically, assumption **(A3)** makes it possible to obtain an *accurate* estimation of the coefficients $\{\gamma_i^*\}_{i=1}^s$. Assumption **A2** ensures that labeled examples are associated with *all* the top eigenfunctions, and therefore a reliable estimation can be obtained for *all* the coefficients through the regression analysis. Intuitively, assumption **(A2)** ensures that $|\phi_i(\mathbf{x}_j)|, j \in [n]$ on the labeled examples are not zeros, which is due to $\mathrm{E}[\phi_i(\mathbf{x})]$ is fixed and $\max_{\mathbf{x}} |\phi_i(\mathbf{x})|$ is small, otherwise we cannot obtain an accurate estimation of $\gamma^*$. Actually, it is notable that we only need to bound the first $s$ eigenfunctions in $M(s) = \max_{\mathbf{x}} \sum_{i=1}^s \phi_i^2(\mathbf{x})$, a key quantity in Proposition 2. From another point of view, if we bound $\max_{\mathbf{x}} |\phi_i(\mathbf{x})| \leq \|\phi_i\|_{\mathcal{H}_\kappa} = 1/\sqrt{\lambda_i}$ (Smale & Zhou, 2009, pg. 9), then if the first $s$ eigenvalues are large, we can expect the maximum value of the first $s$ eigenfuncitons is small. An example satisfying this property is the Sobolev space of functions defined on the domain $[0,1]^d$ with uniform distribution (see (Koltchinskii, 2011, pg. 16)).

The following theorem shows the generalization error of Algorithm 1 for an infinite number of unlabeled examples provided that assumptions **(A0∼A3)** hold.

**Theorem 1.** *Assume (**A0** ∼ **A3**) hold. Set* $s = (aR/\varepsilon)^{2/(p-1)}$. *Then, with a probability* $1 - 2N^{-3}$, *we have*

$$\mathrm{E}_{\mathbf{x}}\left[(\widehat{g}(\mathbf{x}) - f(\mathbf{x}))^2\right] \leq O(\varepsilon^2),$$

*where* $\widehat{g}(\cdot)$ *is the function learned by Algorithm 1.*

**Remark 3** According to (2), $\varepsilon^2$ is the optimal regression error that can be achieved by a prediction function in $\mathcal{H}_\kappa$. Hence, Theorem 1 shows that given an infinite number of unlabeled examples, the prediction function learned by Algorithm 1 achieves almost the optimal performance (up to a constant).

**Remark 4** It is also useful to compare the bound in Theorem 1 to the generalization error bound of supervised learning. According to (Tsybakov, 2008), the minimax optimal error if supervised regression (i.e., the best possible regression error of the worst possible distribution) is bounded by $\Omega(n^{-p/(p+1)})$ [2]. So if we take the value in assumption (**A3**) for $n \propto \epsilon^{-4/(p-1)}$, then the generalization error for supervised regression is $\Omega(\varepsilon^{4p/(p^2-1)})$. Compared to our bound (i.e., $O(\varepsilon^2)$), when $p > 1 + \sqrt{2}$, we have $4p/(p^2-1) < 2$, implying that the generalization error bound of Algorithm 1 is better than that for supervised regression.

### 3.2. Generalization error for a finite number of unlabeled examples

We now consider the scenario where only a finite number (i.e., $N$) of unlabeled examples are available. The key challenge arising from the finite sample analysis is that we do not have access to the eigenfunctions and eigenvalues of $L$. Instead, we have to approximate the eigenfunctions and eigenvalues of $L$ by its empirical counterpart $\widehat{L}_N$. These approximation errors make the analysis more involved. To ensure that the approximation does not significantly increase the regression error, we make the following assumptions:

- **B1** *Skewed eigenvalue distribution of* $\widehat{L}_N$. We assume eigenvalues $\widehat{\lambda}_i, i = 1, 2, \ldots$ follow a power law distribution, i.e., there exists a small constant $a$ and power index $p > 2$, such that
$$\widehat{\lambda}_k \leq a^2 k^{-p}, k = 1, 2, \ldots.$$
- **B2** *Bounded eigenfunctions.* There exists a small constant $\widehat{C}$ such that $\max_{\mathbf{x}\in\mathcal{X}} \max_i |\widehat{\phi}_i(\mathbf{x})/\sqrt{\lambda_i}| \leq \widehat{C}$.
- **B3** *Sufficient number of labeled examples.* We require the number of labeled examples to be larger than $n_0$ where $n_0$ is defined as
$$n_0 = 64\widehat{C}^2 \ln^2(2N^3)\left(\frac{Ra}{\varepsilon}\right)^{4/(p-1)}.$$
- **B4** *Sufficiently large eigengap.* Let $r_s = \lambda_s - \lambda_{s+1}$ be the gap between the $s$-th eigenvalue and $(s+1)$-th eigenvalue of $L$. We assume the eigengap $r_s$ is sufficiently large for $s = (Ra/\varepsilon)^{2/(p-1)}$, i.e., $r_s \geq 3\tau_N^{2/3}$, where $\tau_N = \frac{12 \ln N}{\sqrt{N}}$.

**Remark 5** Assumptions (**B1∼B3**) are the "empirical" versions of assumptions (**A1∼A3**). Note that unlike assumption (**A2**) where $|\phi_i(\mathbf{x})|$ is assumed to be

[2] We use $\Omega(\cdot)$, instead of $O(\cdot)$, since it is a minimax optimal bound.

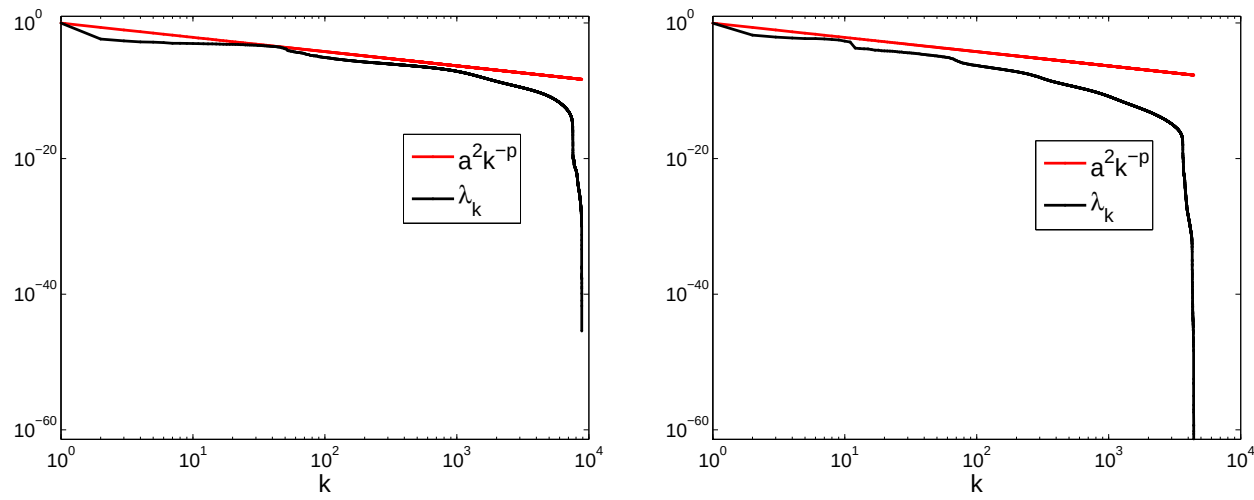

(a) insurance (a=0.98) (b) wine (a=0.97)

*Figure 1.* Eigenvalue distribution for the insurance and wine datasets

bounded, in assumption (**B2**), we assume $|\widehat{\phi}_i(\mathbf{x})/\sqrt{\lambda_i}|$ to be bounded. This is because $\phi_i(\mathbf{x})$ is normalized with respect to the distribution $P_{\mathcal{X}}$, while $\widehat{\phi}_i(\mathbf{x})$ is normalized with respect to the functional norm since the marginal distribution $P_{\mathcal{X}}$ is unknown. The most important feature of the finite sample analysis is that we introduce a new assumption (**B4**), where the number of unlabeled examples $N$ plays an important role to bound the eigengap. This additional assumption is designed to address the approximation error in replacing the eigenfunctions of $L$ with the eigenfunctions of $\widehat{L}_N$.

**Theorem 2.** *Assume (**A0**) and (**B1**~**B3**) hold. Set $s = (aR/\varepsilon)^{2/(p-1)}$, and assume*

$$N \geq \max\left(144R^2[\ln N]^2 r_s^{-2}\varepsilon^{-2}, 144R^4a^2[\ln N]^2\varepsilon^{-4}\right).$$

*Then, with a probability $1 - 4N^{-3}$, we have*

$$\mathrm{E}_{\mathbf{x}}[(\widehat{g}(\mathbf{x}) - f(\mathbf{x}))^2] \leq O(\varepsilon^2).$$

As indicated by Theorem 2, the prediction function learned by Algorithm 1 achieves almost the optimal regression error (up to a constant) provided that all the assumptions hold and the number of unlabeled examples is sufficiently large.

Finally, to partially verify the assumptions, we examine the eigenvalue distributions for the chosen datasets (described in Section 2.2), as shown in Figure 1. Due to space limitation, we put the figure for the temperature dataset in the supplementary material. We also show in Figure 1 the curves of $a^2k^{-p}$ with $p = 2.1$. It is very clear that the eigenvalues follow a skewed distribution with the power index $p > 2$.

## 4. Analysis

We present the full analysis for the case of infinite number of unlabeled examples, and only sketch the analysis for finite number of unlabeled examples due to lack of space. More detailed analysis can be found in the supplementary materials.

### 4.1. Analysis for an infinite number of unlabeled examples

When we have an infinite number of unlabeled examples, the learned prediction function is given by $\widehat{g}(\mathbf{x}) = \sum_{j=1}^{s}\gamma_j^*\phi_j(\mathbf{x})$, where $\gamma^* = (\gamma_1, \cdots, \gamma_s^*)^\top$ is obtained by solving the following optimization problem:

$$\gamma^* = \arg\min_{\gamma}\left[\mathcal{L}(\gamma) = \sum_{i=1}^{n}\left(\sum_{j=1}^{s}\gamma_j\phi_j(\mathbf{x}_i) - f(\mathbf{x}_i)\right)^2\right]. \tag{7}$$

Using the eigenfunctions of $L$, we write $g(\mathbf{x})$, the optimal prediction function defined in (3), as $g(\mathbf{x}) = \sum_j \alpha_j\phi_j(\mathbf{x})$. We define $g_s(\mathbf{x})$, the projection of $g(\mathbf{x})$ into the subspace spanned by the top $s$ eigenfunctions, as

$$g_s(\mathbf{x}) = \sum_{j=1}^{s}\alpha_j\phi_j(\mathbf{x}).$$

Using $g_s(\mathbf{x})$, we decompose the generalization error of $\widehat{g}(\mathbf{x})$ into two parts, i.e.,

$$\begin{aligned}&\mathrm{E}_{\mathbf{x}}[(\widehat{g}(\mathbf{x}) - f(\mathbf{x}))^2]\\ &\quad\leq\ 2\mathrm{E}_{\mathbf{x}}[(\widehat{g}(\mathbf{x}) - g_s(\mathbf{x}))^2] + 2\mathrm{E}_{\mathbf{x}}[(g_s(\mathbf{x}) - f(\mathbf{x}))^2].\end{aligned}$$

The following lemmas bound the two terms on the R.H.S. of the above inequality, separately.

**Lemma 1.** *Under assumption* (**A1**), *for any $s \geq 1$, we have*

$$\mathrm{E}_{\mathbf{x}}\left[(g_s(\mathbf{x}) - f(\mathbf{x}))^2\right] \leq 2\varepsilon^2 + \frac{2a^2R^2}{s^{p-1}} \triangleq \varepsilon_s^2.$$

**Lemma 2.** *Under assumptions* (**A2**~**A3**) *and $s = (aR/\epsilon)^{2/(p-1)}$, with a probability at least $1 - 2N^{-3}$, we have*

$$\mathrm{E}_{\mathbf{x}}\left[(\widehat{g}(\mathbf{x}) - g_s(\mathbf{x}))^2\right] \leq 2\eta^2,$$

*where* $\eta^2 = 2\left(\varepsilon_s^2 + 2\varepsilon_s\varepsilon_{\max}\sqrt{\frac{3\ln N}{n}} + \frac{\varepsilon_{\max}^2\ln N}{n}\right)$.

As indicated by Lemma 1, assumption (**A1**) guarantees an additional small regression error when constraining the solution to the subspace spanned by the top eigenfunctions of $L$. As indicated by Lemma 2, assumptions (**A2**~**A3**) ensure that $g_s(\mathbf{x})$, the projection of $g(\mathbf{x})$ into the subspace spanned by the top eigenfunctions, can be accurately estimated from the labeled examples. It is easy to see that Theorem 1 immediately follows Lemma 1 and Lemma 2 by noting that $\varepsilon_s^2 = O(\varepsilon^2)$ and $\eta^2 = O(\varepsilon^2)$ when we set $s = (Ra/\varepsilon)^{2/(p-1)}$. Below, we show how to prove both lemmas.

**Proof of Lemma 1** We first show that $\sum_{i=s+1}^{\infty}\alpha_i^2$ is bounded. Since $\|g\|_{\mathcal{H}_\kappa}\leq R$, we have

$$R^2 \geq \langle g, g\rangle_{\mathcal{H}_\kappa} = \sum_{i=1}^{\infty}\alpha_i^2\|\phi_i\|^2_{\mathcal{H}_\kappa} = \sum_{i=1}^{\infty}\frac{\alpha_i^2}{\lambda_i},$$

and therefore

$$\sum_{i=s+1}^{\infty}\alpha_i^2 \leq R^2\sum_{i=s+1}^{+\infty}\lambda_i \leq \frac{a^2R^2}{(p-1)s^{p-1}} \leq \frac{a^2R^2}{s^{p-1}}.$$

Then we bound the regression error of $g_s(\mathbf{x})$ as follows:

$$\begin{aligned}\mathrm{E}_{\mathbf{x}}\left[(g_s(\mathbf{x})-f(\mathbf{x}))^2\right] &\leq 2\mathrm{E}_{\mathbf{x}}\left[(g(\mathbf{x})-f(\mathbf{x}))^2\right] \\ &\quad +2\mathrm{E}_{\mathbf{x}}\left[\sum_{i,j=s+1}^{\infty}\alpha_i\alpha_j\phi_i(\mathbf{x})\phi_j(\mathbf{x})\right] \\ &= 2\varepsilon^2 + 2\sum_{i=s+1}^{\infty}\alpha_i^2 \leq 2\varepsilon^2 + \frac{2a^2R^2}{s^{p-1}} \triangleq \varepsilon_s^2.\end{aligned}$$

**Proof of Lemma 2** The proof of Lemma 2 is significantly more involved. We first introduce some notations. Let $\mathbf{z}_i = (\phi_1(\mathbf{x}_i),\ldots,\phi_s(\mathbf{x}_i))^\top$ be the vector representation of $\mathbf{x}_i$ derived from the first $s$ eigenfunctions. Let $Z = (\mathbf{z}_1,\ldots,\mathbf{z}_n)$ include the representations of all labeled examples, and let $\mathbf{y}_l = (f(\mathbf{x}_1),\ldots,f(\mathbf{x}_n))^\top$. Using $Z$, we rewrite $\mathcal{L}(\gamma)$ in (7) as

$$\mathcal{L}(\gamma) = \gamma^\top ZZ^\top\gamma - 2\gamma^\top Z\mathbf{y}_l + \|\mathbf{y}_l\|_2^2.$$

The following proposition bounds $\mathrm{E}_{\mathbf{x}}[(\widehat{g}(\mathbf{x})-g_s(\mathbf{x}))^2]$ using the minimum eigenvalue of $ZZ^\top$.

**Proposition 1.** *Assume $ZZ^\top$ is nonsingular. With a probability at least $1-N^{-3}$, we have*

$$\mathrm{E}_{\mathbf{x}}\left[(\widehat{g}(\mathbf{x})-g_s(\mathbf{x}))^2\right] = \|\alpha^s-\gamma^*\|_2^2 \leq \frac{n\eta^2}{\lambda_{\min}(ZZ^\top)}.$$

The following proposition bounds the minimum eigenvalue of $ZZ^\top$.

**Proposition 2.** *With a probability at least $1-N^{-3}$, where $N>0$ is a large number, we have*

$$\frac{1}{n}\lambda_{\min}(ZZ^\top) \geq 1 - \frac{4M(s)\ln(2N^3)}{\sqrt{n}},$$

*where $M(s) = \max_{\mathbf{x}\in\mathcal{X}}\sum_{i=1}^{s}\phi_i^2(\mathbf{x})$.*

The proof for Proposition 1 and 2 can be found in the supplementary materials. Now we are ready to prove Lemma 2.

According to assumptions **A2**$\sim$**A3** and Proposition 2, we have, with a probability at least $1-N^{-3}$

$$\frac{1}{n}\lambda_{\min}(ZZ^\top) \geq 1 - \frac{4M(s)\ln(2N^3)}{\sqrt{n}} \geq \frac{1}{2}.$$

Combining the above inequality with Proposition 1, we have, with a probability at least $1-2N^{-3}$,

$$\mathrm{E}_{\mathbf{x}}\left[(\widehat{g}(\mathbf{x})-g_s(\mathbf{x}))^2\right] \leq 2\eta^2.$$

### 4.2. Analysis for a finite number of unlabeled examples

Define $\gamma^*$ the optimal solution that minimizes the regression error using the eigenfunctions of $\widehat{L}_N$, i.e.,

$$\gamma^* = \arg\min_{\gamma\in\mathbb{R}^s}\sum_{i=1}^{n}\left(f(\mathbf{x}_i) - \sum_{k=1}^{s}\gamma_k\widehat{\phi}_k(\mathbf{x}_i)\right)^2.$$

We further define $\widehat{\gamma}_i^* = \gamma_i^*\sqrt{\lambda_i}, i=1,\cdots,s$, and write $\widehat{g}(\mathbf{x})$ learned in the presence of a finite number of unlabeled examples as $\widehat{g}(\mathbf{x}) = \sum_{i=1}^{s}\widehat{\gamma}_i^*\frac{\widehat{\phi}_i(\mathbf{x})}{\sqrt{\lambda_i}}$. We also introduce $h_s(\mathbf{x})$ as follows

$$h_s(\mathbf{x}) = \sum_{i=1}^{s}\alpha_i\frac{\widehat{\phi}_i(\mathbf{x})}{\sqrt{\lambda_i}}.$$

where $\{\alpha_i\}_{i=1}^s$ are the coefficients defined in $g(\mathbf{x})$. Similar to the previous analysis, we bound the generalization error of $\widehat{g}(\mathbf{x})$ by

$$\begin{aligned}&\mathrm{E}_{\mathbf{x}}[(\widehat{g}(\mathbf{x})-f(\mathbf{x}))^2] \\ &\leq 2\mathrm{E}_{\mathbf{x}}[(\widehat{g}(\mathbf{x})-h_s(\mathbf{x}))^2] + 2\mathrm{E}_{\mathbf{x}}[(h_s(\mathbf{x})-f(\mathbf{x}))^2].\end{aligned}$$

We follow the same path as in the infinite case and present two lemmas to bound the two terms on R.H.S. of the above inequality.

**Lemma 3.** *Under assumptions* **B1, B3** *and $N \geq 144s^{2p-2}[\ln N]^2a^{-2}$, with a probability at least $1-2N^{-3}$, we have*

$$\mathrm{E}_{\mathbf{x}}[(h_s(\mathbf{x})-f(\mathbf{x}))^2] \leq 4\varepsilon_s^2 + \frac{36R^2\tau_N^2}{r_s^2} \triangleq \widehat{\varepsilon}_s^2.$$

**Lemma 4.** *Under assumptions* **B1**$\sim$**B3***, with a probability at least $1-4N^{-3}$, we have*

$$\mathrm{E}_{\mathbf{x}}\left[(\widehat{g}(\mathbf{x})-h_s(\mathbf{x}))^2\right] \leq 4\widehat{\eta}^2.$$

*where* $\widehat{\eta}^2 = 2\left(\widehat{\varepsilon}_s^2 + 2\widehat{\varepsilon}_s\varepsilon_{\max}\sqrt{\frac{3\ln N}{n}} + \frac{\varepsilon^2_{\max}\ln N}{n}\right).$

The proof for Lemma 4 and Lemma 3 can be found in the supplementary materials.

*Proof of Theorem 2.* Using the condition $N \geq 144R^2[\ln N]^2/[r_s^2\varepsilon^2]$, we have $36R^2\tau_N^2/r_s^2 \leq O(\varepsilon^2)$. When we set $s = (Ra/\varepsilon)^{2/(p-1)}$, we have $\varepsilon_s^2 = O(\varepsilon^2)$, $\widehat{\varepsilon}_s^2 = O(\varepsilon^2)$ and $\widehat{\eta} = O(\varepsilon^2)$ . By Lemma 3 and Lemma 4, we have, with a probability $1-4N^{-3}$,

$$\mathrm{E}_{\mathbf{x}}[(\widehat{g}(\mathbf{x})-f(\mathbf{x}))^2] \leq 2\widehat{\varepsilon}_s^2 + 8\widehat{\eta}^2 = O(\varepsilon^2).$$

□

## 5. Conclusions

In this work, we present a very simple algorithm for semi-supervised learning. Our analysis shows that under appropriate assumptions about the integral operator, the proposed algorithm achieves a better generalization error than a supervised learning algorithm. In the future, we plan to further improve the scalability of the proposed algorithm by exploring different approaches (e.g., the Nyström method) for efficiently estimating eigenfunctions from a large number of unlabeled examples.

## Acknowledgments

The work was supported in part by the U.S. Army Research Laboratory under Cooperative Agreement No. W911NF-09-2-0053 (NS-CTA), NSF IIS-0905215, NSF IIS-0643494, U.S. Air Force Office of Scientific Research MURI award FA9550-08-1-0265, and Office of Navy Research (ONR Award N00014-09-1-0663 and N00014-12-1-0431). The views and conclusions contained in this paper are those of the authors and should not be interpreted as representing any funding agencies.